\documentclass[runningheads]{llncs}
\usepackage[T1]{fontenc}
\usepackage{graphicx}
\usepackage{multirow}
\usepackage{amsmath}
\usepackage{subcaption}
\usepackage{booktabs}
\usepackage{amsmath}
\usepackage{amsfonts}
\usepackage{xcolor}
\usepackage{etoolbox}
\usepackage{cite}
\newbool{DRAFT}
\setbool{DRAFT}{true}
\newcommand{\COMMENT}[3]{\ifbool{DRAFT}{\textcolor{#1}{[#2 -#3]}}{}}

\begin{document}
\newcommand{\modelname}[1]{DeformGS{#1}}

\title{\modelname{}: Scene Flow in Highly Deformable Scenes for Deformable Object Manipulation  }
\titlerunning{DeformGS}
% \author{Bardienus P. Duisterhof\inst{1}\orcidID{0000-0002-2797-4898} \and
% Mandi Zhao\inst{2}\orcidID{0009-0005-5877-920X} \and
% Yunchao Yao\inst{1}\orcidID{0009-0008-6537-7075}  \and
% Jia-Wei Liu\inst{4}\orcidID{0000-0002-9766-2419} \and
% Jenny Seidenschwarz\inst{1,5}\orcidID{0000-0002-8955-0767} \and
% Mike Zheng Shou\inst{4}\orcidID{0000-0002-7681-2166} \and
% Deva Ramanan\inst{1}\orcidID{0009-0008-9180-8983} \and
% Shuran Song\inst{2}\orcidID{0000-0002-8768-7356} \and
% Stan Birchfield\inst{3}\orcidID{1111-2222-3333-4444} \and
% Bowen Wen\inst{3}\orcidID{0000-0002-9207-6103} \and
% Jeffrey Ichnowski\inst{1}\orcidID{0000-0003-4874-9478}
% }
\author{Bardienus P. Duisterhof\inst{1} \and
Mandi Zhao\inst{2} \and
Yunchao Yao\inst{1}  \and
Jia-Wei Liu\inst{4} \and \\
Jenny Seidenschwarz\inst{1,5} \and
Mike Zheng Shou\inst{4} \and
Deva Ramanan\inst{1} \and
Shuran Song\inst{2} \and
Stan Birchfield\inst{3} \and
Bowen Wen\inst{3} \and
Jeffrey Ichnowski\inst{1}
}
\authorrunning{B.P. Duisterhof et al.}
\institute{Carnegie Mellon University, The Robotics Institute \email{\{bduister,jeffi\}@cmu.edu} \and
Stanford University \and
NVIDIA \and 
National University of Singapore \and 
Technical University of Munich
}
\maketitle              % typeset the header of the contribution
%
% \begin{abstract}
% The abstract should briefly summarize the contents of the paper in

% \keywords{First keyword  \and Second keyword \and Another keyword.}
% \end{abstract}
%
%
%
\begin{abstract}
% \textcolor{red}{Old abstract} 
Teaching robots to fold, drape, or reposition deformable objects such as cloth will unlock a variety of automation applications.
%enable new applications in industry. %Recent work has demonstrated robust manipulation of rigid objects using fast and robust 6D pose estimation enabled by foundation models. 
While remarkable progress has been made for rigid object manipulation, 
manipulating deformable objects poses unique challenges, including frequent occlusions, infinite-dimensional state spaces
%beyond 6D
and complex dynamics. Just as object pose estimation and tracking have aided robots for rigid manipulation, dense 3D tracking (scene flow) of highly deformable objects will enable new applications in robotics while aiding existing approaches, such as imitation learning or creating digital twins with real2sim transfer.
% Let's avoid paragraphs in abstracts
We propose \modelname{}, an approach to recover scene flow in highly deformable scenes, using simultaneous video captures of a dynamic scene from multiple cameras. 
\modelname{} builds on recent advances in Gaussian splatting, a method that learns the properties of a large number of Gaussians for state-of-the-art and fast novel-view synthesis. \modelname{} learns a deformation function to project a set of Gaussians with canonical properties into world space. The deformation function uses a neural-voxel encoding and a multilayer perceptron (MLP) to infer Gaussian position, rotation, and a shadow scalar. 
% Let's avoid paragraphs in abstracts
We enforce physics-inspired regularization terms based on conservation of momentum and isometry, which leads to trajectories with smaller trajectory errors. We also leverage existing foundation models SAM and XMEM to produce noisy masks, and learn a per-Gaussian mask for better physics-inspired regularization. \modelname{} achieves high-quality 3D tracking on highly deformable scenes with shadows and occlusions. In experiments, \modelname{} improves 3D tracking by an average of 55.8 \% compared to the state-of-the-art. With sufficient texture, % such as in scene 6, 
\modelname{} achieves a median tracking error of 3.3 mm on a cloth of 1.5 $\times$ 1.5 m in area. Website: \url{https://deformgs.github.io}
\end{abstract}

%Bart: high-level points to make
% 1): Robust and fast deformable object manipulation is useful
% 2): Recent advances in pose estimation have unlocked new applications in rigid object manipulation, but are insufficient to recover a deformable state

% Old motivation
% Accurate 3D tracking in highly deformable scenes with occlusions and shadows can facilitate new applications in robotics, augmented reality, and generative AI. However, tracking under these conditions is extremely challenging due to the ambiguity that arises with large deformations, shadows, and occlusions.    
\section{Introduction}
\label{sec:intro}
\begin{figure}[t]
    \centering
    \includegraphics[width=0.8\linewidth]{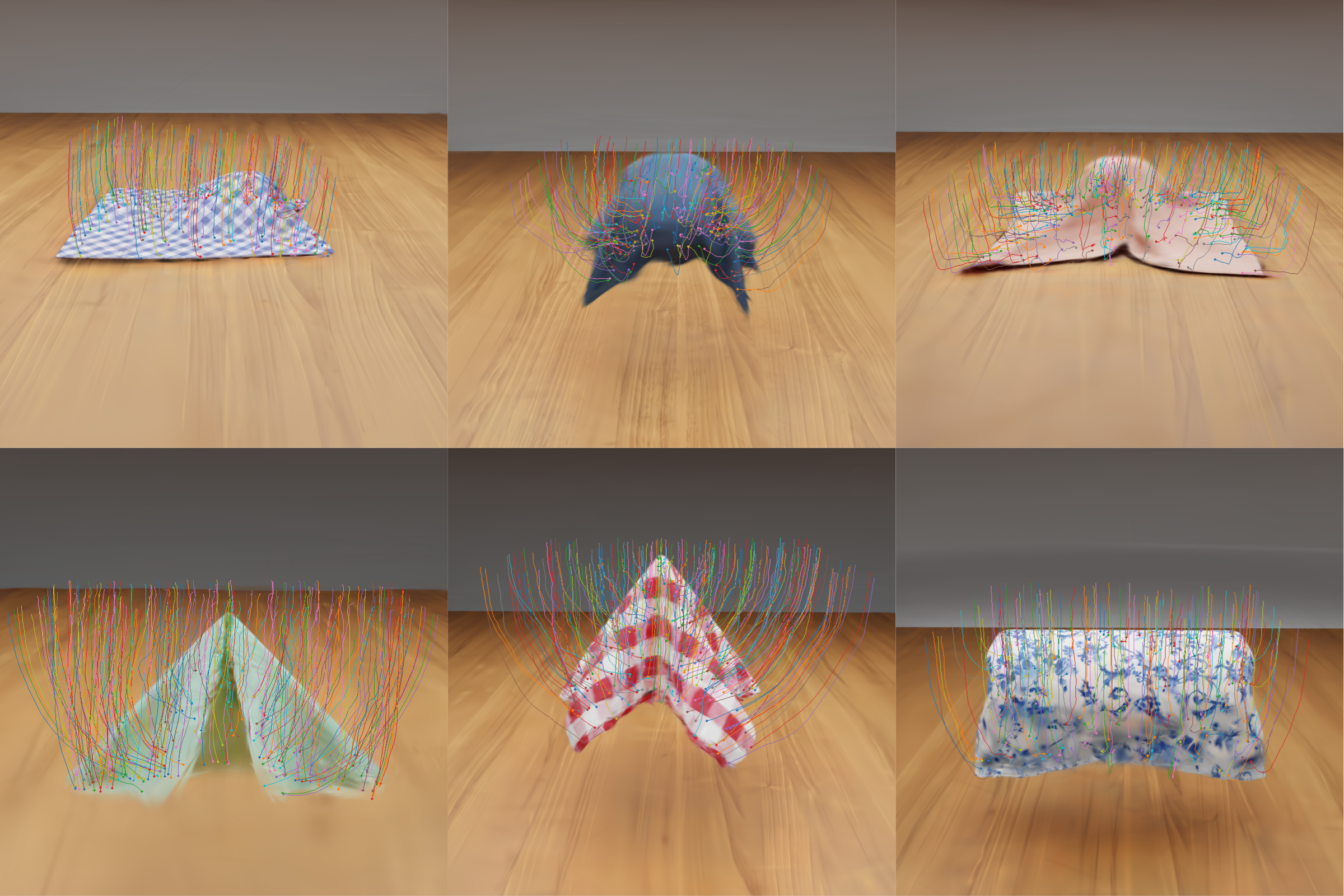}
    \caption{We propose \modelname{}, a method that improves state-of-the-art methods for accurate 3D point tracking in highly deformable scenes. This figure shows the rendering and tracking of \modelname{} in the six dynamic Blender~\cite{blender} scenes used for evaluation. We will refer to the scenes in this Figure as Scenes 1, 2, 3, 4, 5 and 6 ordered from left to right.}
    \label{fig:teaser}
\end{figure}

% pitch why cloth tracking would be useful
Recent advances in robot learning have demonstrated impressive performance on challenging tasks, including rigid and deformable object manipulation. Scaling these approaches to deployment will require an improvement in robustness and learning from few demonstrations.
A promising avenue for improving in robot learning performance are intermediate representations and foundation models, including
%Intermediate representations enabled by foundation models have shown to improve robotic task performance.
%This includes
6D object pose estimation~\cite{deng2020self,devgon2020orienting,lin2024sam,taher2024fit,wen2020se,wen2024foundationpose,xiang2017posecnn}, semantic latent features~\cite{oquab2024dinov2}, and 2D pixel-wise tracking~\cite{karaev2023cotracker,wen2023anypoint}. 
However, perception and representations that will lead to robust manipulation of deformable objects remains an open challenge, due to self-occlusions, shadows, and varying (or lack of) textures. 

% Generative AI and augmented reality applications can benefit from novel view synthesis for training, interaction, and scene editing~\cite{instructnerf2023}. Robots can benefit from tracking task-relevant geometries to dexterously manipulate deformable objects~\cite{seita_fabrics_2020,10.1007/978-3-031-25555-7_4,10160574} and use novel-view synthesis to augment learning~\cite{haarnoja2023learning}. Simulation methods can improve fidelity using \emph{real2sim} and related approaches that leverage real-world tracking data for parameter and method tuning~\cite{9811651}. However, tracking moving and morphing points poses many non-trivial challenges: points change color as they pass through shadows, occlusions interfere with the continuous visibility of points, and internal and external forces, such as material tension and air currents, cause points to move unpredictably.

% paragraph 2: introduction related work
% TODO: add robotics angle
Three-dimensional dense point tracking, or \emph{3D scene flow}, is a useful representation for robot manipulation, as it provides flexibility to represent high-dimensional dynamic state changes, while the deformable objects drops, deforms, and drapes during manipulation. In particular, dense 3D scene flow can be an input to imitation learning policies~\cite{wen2023anypoint,bharadhwaj2024track2act}, can be used to learn a transition model~\cite{shi2022robocraft}, to identify and track task-relevant key points, or to create a digital twin through real2sim transfer. Recent work in monocular tracking has seen improvements in performance on datasets such as TAP-Vid~\cite{doersch2023tapvid}, but it remains unclear how to effectively lift from 2D tracking to 3D for robotic spatial understanding in challenging highly deformable scenes.
%although it remains unclear how suitable foundation tracking models are for inferring scene flow of highly deformable scenes. Deformable objects pose unique challenges, as occlusion, textureless surfaces, and unobserved physics are common.

To overcome these limitations, Gaussian Splatting provides a promising avenue. Recent work demonstrated Gaussian Splatting~\cite{kerbl3Dgaussians,keselman2022fuzzy} can yield state-of-the-art novel-view synthesis and rendering speeds exceeding 100\,fps. Concretely, 3D Gaussian Splatting uses a fast differentiable renderer to fit the colors, positions, and covariances of a set of Gaussians. An extension of 3D Gaussian splatting~\cite{dyna3dgs} showed dynamic scenes can be modeled by explicitly optimizing the properties of Gaussians over time, resulting in novel-view synthesis and scene flow.

Explicitly optimizing the Gaussian pose as in Dynamic 3D Gaussians~\cite{dyna3dgs} may result in degraded performance with large deformations and shadows. The Gaussian properties may converge to local optima, especially in scenes with large deformations, strong shadows, and occlusions.

% paragraph 3: Our approach part I - method
We propose \modelname{}, a method that uses time-synchronized image frames from a calibrated multi-camera setup to track 3D geometries of deformable objects as they move through shadows and occlusions. \modelname{} learns the canonical state of a set of Gaussians and a deformation function that maps the Gaussians into world space. This enables tracking by recovering scene flow, and novel-view rendering (through splatting) using a fast differentiable rasterizer.

% paragraph 4: Our approach part II - results
We evaluate \modelname{} in six photo-realistic synthetic scenes of varying difficulty.  The scenes contain large deformations, shadows, and occlusions (Figure~\ref{fig:teaser} shows the scenes and tracking trajectories computed by \modelname{}). Empirical results show that \modelname{} infers 55.8\,\% more accurate 3D tracking results compared to previous state-of-the-art~\cite{4dgs,dyna3dgs}. In a scene with a 1.5 m $\times$ 1.5 m cloth (i.e., Scene 1), \modelname{} can track cloth deformation with as low as 3.3\,mm median tracking error. 

% To the best of our knowledge, \modelname{} is the first approach to achieve tracking at this accuracy. 

We also evaluate \modelname{} in the real world on the Robo360~\cite{liang2023robo360} dataset. We show qualitative results for tracking rigid and deformable objects in cluttered scenes, and study two robotics applications: (1) real2sim transfer to create a digital twin, and (2) tracking task-relevant keypoints for downstream grasping applications.

% paragraph 5: Contributions summary 
In summary, our contributions are as follows:
\begin{itemize}
    \item We provide the first approach designed to accurately perform 3D dense tracking for deformable objects using 4D Gaussians.
    \item We provide experiments that suggest state-of-the-art performance in simultaneous 3D metric tracking and novel view synthesis. \modelname{} improves tracking accuracy by an average of 55.8\% in synthetic experiments and demonstrates robust 3D tracking in the real world for deformable objects. The latter can be exploited as a representation for imitation learning and represents a new method for building digital twins.    
    \item A set of six synthetic scenes with large deformations, strong shadows, and occlusions. We will open-source the scenes and as well as the source code. 
\end{itemize}

% \JI{Some reviewers push back on the `experiments' as a contribution. Let's update this contribution to make a statement about how \modelname{} is useful/applicable.}

% \input{sec/2_formatting}
% \input{sec/3_finalcopy}

\section{Related Work}
\label{sec:related}

% \modelname{} builds on prior work in novel view synthesis in dynamic scenes to enable simultaneous 3D tracking and novel view synthesis. We discuss prior works in novel view synthesis in static and dynamic scenes, as well as methods for point tracking. 

\subsection{Neural Rendering for Novel View Synthesis}
\modelname{} builds on prior work in novel-view synthesis, and uses photometric consistency as a signal to achieve 3D tracking. A popular novel view synthesis approach is NeRF~\cite{mildenhall2020nerf}. It uses neural networks to learn scene representations that are capable of photo-realistic novel view reconstruction. Particle-based methods use a more explicit representation than typical NeRF-based approaches. \modelname{} builds on 3D Gaussian Splatting \cite{kerbl3Dgaussians,keselman2022fuzzy} which belongs to the latter caregory. \cite{kerbl3Dgaussians} proposed a differential rasterizer to render a large number of Gaussian `splats,' each with their state including color, position, and covariance matrix. Contrary to the NeRF-based approaches, Gaussian splatting achieves real-time rendering of novel views with state-of-the-art performance.

% Subsequent works improve NeRF along several axes: one is to speed up training and inference time via novel representations and system optimizations~\cite{mueller2022instant,DBLP:journals/corr/abs-2103-13744,liu2020neural,yu2021plenoctrees,SunSC22,9710021,10.1145/3450626.3459863,mueller2022instant,mubarik2023hardware}, or depth supervision~\cite{9880067,attal2021torf,wei2021nerfingmvs,neff2021donerf,Sucar:etal:ICCV2021}. Other works extend NeRF to more challenging conditions, such as sparser camera views~\cite{zhang2021ners,SRF,sparf2023,Niemeyer2021Regnerf}, fewer extrinsic camera calibrations~\cite{lin2021barf,yen2020inerf,chen2023local,Jeong_2021_ICCV}, transparent objects~\cite{pmlr-v205-kerr23a,IchnowskiAvigal2021DexNeRF} and reflective surfaces~\cite{verbin2022refnerf}. 

\subsection{Dynamic Novel View Synthesis} 
The assumption of static scenes in neural rendering approaches
prevents application to real-world scenarios with moving objects or humans, such as the dynamic and deformable scenes in this work. One line of work to address this assumption is adding a time dimension to NeRF modeling~\cite{du2021nerflow,Gao-ICCV-DynNeRF,9578364,9577376}. Prior works either condition the neural field on explicit time input or a time embedding. Another line of work learns a deformation field to map 4D points into a canonical space~\cite{pumarola2020d,park2021nerfies}, i.e., every 4D point in space and time maps to a 3D point in a canonical NeRF.  DeVRF~\cite{liu2022devrf} proposed to model the 3D canonical space and 4D deformation field of a dynamic, non-rigid scene with explicit and discrete voxel-based representations. 

% It achieves convergence of optimizing dynamic voxel radiance fields within $10$ minutes on a single NVIDIA GeForce RTX3090 GPU. However, it struggles to reconstruct dynamic scenes with complex deformations and shadows, especially with dynamic occlusions. 

Several recent works extend the above approaches to 3D Gaussian splatting. Dynamic 3D Gaussians~\cite{dyna3dgs} explicitly model the position and covariance matrix of each Gaussian at each time step. This method struggles in dynamic scenes with large deformations, strong shadows, or occlusions. We build on another recent work, 4D Gaussian splatting~\cite{4dgs}, which uses feature encoding techniques proposed in HexPlanes~\cite{cao2023hexplane} and K-planes~\cite{kplane}, and learns a deformation field instead.

% To the best of our knowledge, all prior deformation-based work learned deformations in a non-metric canonical space without reconstructing metric deformations. We propose a method to learn metric deformations from a set of Gaussians with canonical properties, deformed to metric space using a neural-voxel deformation function. By training on adjacent timesteps, \modelname{} allows enforcing physics-informed regularization terms to improve performance beyond photometric consistency.  

\subsection{Point Tracking}
Point tracking methods, usually trained on large amounts of data, aided previous 3D tracking approaches by providing a strong prior~\cite{liu2022devrf}. We also construct several baselines that include point tracking methods (Section~\ref{sec:experiments}). Prior work on point tracking often studies tracking 2D points across video frames, where a dominant approach is training models on large-scale synthetic datasets containing ground-truth point trajectories~\cite{tapvid,doersch2023tapir,pointodyssey,harley2022particlevideo} or dense optical flows~\cite{trackeverything}. Optical flow~\cite{opticalflow,teed2020raft} or scene flow~\cite{sceneflow,Basha2010MultiviewSF,Vogel20153DSF,Guo_2023_ICCV} can also be viewed as single-step point-tracking in 2D and 3D, respectively. 

Another relevant line of work tightly couples dynamic scene reconstruction and motion estimation of non-rigid objects. A predominant setup is fusing RGBD frames from videos of dynamic scenes or objects~\cite{dynamicfusion}. Tracking or correspondence-matching methods see a progression from template-based tracking of objects with known shape or kinematics priors (such as human hand, face or body poses) \cite{Oikonomidis2011EfficientM3,dart,Cao20133DSR}, to more general shapes or scenes~\cite{Zollhfer2014RealtimeNR,bozic2021neural,bozic2020deepdeform}. The main difference from these works is that we do not use depth input, and perform more rigorous quantitative evaluations on tracking specific points.  

Most related to ours is the more recent methods that obtain tracking from neural scene rendering. DCT-NeRF~\cite{DCTNerf} learns a coordinate-based neural scene representation that outputs continuous 3D trajectories across the whole input sequence. PREF~\cite{uii_eccv22_pref} optimizes a dynamic space-time neural field with self-supervised motion prediction loss. Most recently, Luiten et al.~\cite{dyna3dgs} models dynamic 3D Gaussians explicitly across timestamps to achieve tracking. While our work also leverages 3D Gaussians, in contrary to the explicit modeling in Dynamic 3D Gaussians~\cite{dyna3dgs}, we learn a deformation function that scales much better with video length, and we focus on deformable objects that are more challenging than the ball-throwing videos used in~\cite{dyna3dgs}. 

\subsection{Tracking for Robotics}
A core motivation for studying point tracking is the potential it can unlock for robotics applications: for example, RoboTAP~\cite{vecerik2023robotap} shows pre-trained point-tracking models improve sample efficiency of visual imitation learning. It detects task-relevant keypoints, infers where those points should move to, and computes an action that moves them there. Any-point~\cite{wen2023anypoint} learns to predict keypoint tracks, but conditioned on language inputs. Track2Act~\cite{bharadhwaj2024track2act} builds on Any-point by learning a generalizable zero-shot policy, which only needs a few embodiment-specific demonstrations. 

Rigid-body, or 6D, pose tracking and estimation has a rich history in robotics due to it foundational ability to model the world for a robot to manipulate~\cite{wen2022you,morgan2021vision,deng2020self,devgon2020orienting,lin2024sam,taher2024fit,wen2020se,wen2024foundationpose,xiang2017posecnn}. In this work, we propose a deformable object analog of 6D pose tracking with the aim of extending successes to deformable object manipulation.

%\JI{Low priority, but also add a paragraph on RW on rigid-body pose estimation and tracking from the robotics community---this will help make the connection to robotics. Andrew Davidson, Dieter Fox, Ken Goldberg, Kostas Bekris, Bowen Wen :) (and many more) have work in this area.}

While existing methods leverage 2D tracking, and learn an additional policy to output robot actions, \modelname{} provides a more powerful representation that allows for reasoning directly in 3D, instead of in the 2D image space. 
\vspace{-2mm}
\section{Problem Statement}
%We focus on simultaneous 3D tracking and novel view synthesis of highly deformable scenes. %This section defines the problem statement and objectives in mathematical terms.
Given a set of timed image sequences captured from multiple cameras with known intrinsics and extrinsics, the objective is to learn a model that performs 3D tracking and novel view synthesis. Each image sequence is captured over the same time interval $t \in [0, H]$.

\textbf{3D Tracking} The primary goal is to recover the trajectory of any point in a dynamic scene by modeling the deformation of Gaussians over time. %, and with that dense 3D tracking. 
Thus, the objective is to find a function $x_t = Q(x_0,t_0,t)$, where $x_0 \in \mathbb{R}^3$ is the location of a point of interest at a chosen time $t_0 \in [0,H] $, while $x_t \in \mathbb{R}^3$ is the location of the same point at another chosen time $t \in [0,H]$. %For the methods considered, t
The function $Q$ is valid for any point $x_0$ and any $t \in [0,H]$, allowing for tracking of any point in space. %Dynamic 3D Gaussians~\cite{dyna3dgs} can only be queried at the timesteps available during training such that $ \{t_0, t \} \in \mathcal{T}$ where $\mathcal{T}$ is the set of all times $t$ during training. 

\textbf{Novel View Synthesis} The secondary goal is to achieve accurate scene flow by using photometric consistency as a supervision signal. To achieve this, the objective is to recover novel views from arbitrary viewpoints. 
% The secondary goal is to recover novel views from arbitrary viewpoints by modelling accurate scene flow using photometric consistency as a supervision signal. 
The extrinsics at any viewpoint can be captured by matrix $P$, with $P= K[R|T]$. Here $K$ is the intrinsics matrix, $R$ is the rotation matrix of a camera with respect to the world frame, and $T$ is the translation vector with respect to the world frame. Concretely, the goal is to learn a function $V$ such that $I_{P,t} = V(P,t)$, where $I_{P,t}$ is an image rendered from a camera with extrinsics $P$ at time $t$. As with the tracking objective, the time parameter is valid for any $t \in [0,H]$.%All methods considered can be queried for any matrix $P$, reconstructing any arbitrary novel viewpoints. Again, along the time axis Dynamic 3D Gaussians~\cite{dyna3dgs} can only be queried at the training time steps.

\section{Preliminary}
% This section summarizes prior work on which \modelname{} builds.
% \modelname{} builds on Neural Radiance Fields~\cite{mildenhall2020nerf}, Guassian Splatting~\cite{kerbl3Dgaussians}, and Deformation Fields. This section summaries 

% \subsection{Neural Radiance Fields (NeRF)}
% \JI{Do we need NeRF prelimiaries? If so, this seems insufficient to convey the background on it. Let's consider deleting this, and the ``more explicit scene representation that NeRF'' statement.}
% Neural radiance fields (NeRFs) trace rays through multiple camera views to train an MLP $\Phi_r$
% to infer a radiance (RGB) and density ($\sigma)$ from a world coordinate $(x)$ and view direction $(\theta, \phi)$ inputs, which are typically concatenated with a positional encoding in the form of sinusoids with varying frequencies.
% %Neural radiance fields (NeRFs)~\cite{mildenhall2020nerf} trace rays through the scene, and use an MLP at every point to infer radiance (RGB) and density ($\sigma)$. At every point, the MLP takes in world coordinates $(x,y,z)$ and viewing direction $(\theta,\phi)$ to infer radiance and density using Equation~\ref{eq:nerf_rendering}.
% \begin{equation}
%     \label{eq:nerf_rendering}
%     \Phi_r (x,\theta,\phi) = (RGB \sigma)
% \end{equation}
 
\subsection{Gaussian Splatting}
3D Gaussian Splatting~\cite{kerbl3Dgaussians} deploys an explicit scene representation by rendering a large set of Gaussians each defined by their mean position $\mu$ and covariance matrix $\Sigma$. Given $x \in \mathbb{R}^3$, its Gaussian is
\[%\begin{equation}
\label{eq:gaussian_3d}
    G(x) = e^{-\frac{1}{2}(x-\mu)^T\Sigma^{-1}(x-\mu)},
\]%\end{equation}

Directly optimizing the covariance matrix $\Sigma$ would lead to infeasible covariance matrices, as they must be positive semi-definite to have a physical meaning. Instead, Gaussian Splatting~\cite{kerbl3Dgaussians} proposes to decompose $\Sigma$ into a rotation $R$ and scale $S$ for each Gaussian: % The covariance matrix can then be computed using Equation~\ref{eq:covariance}, and the mean position $\mathcal{X}$ is directly optimized. 
\[%\begin{equation}
\label{eq:covariance}
\Sigma = RSS^TR^T,
\]%\end{equation}
and optimize $R$, $S$, and the mean position.

Given the transformation W of a camera, the covariance matrix can be projected into image space 
%using Equation~\ref{eq:project_gauss}.
as
\[%\begin{equation}
\label{eq:project_gauss}
\Sigma' =  JW\Sigma W^T J^T,
\]%\end{equation}
where $J$ is the Jacobian of the affine approximation of the projective transformation.

% Each pixel is rendered by rendering $\mathcal{N}$ ordered points overlapping the pixel, as given by 
%Equation~\ref{eq:gauss_color}.
During rendering, we compute the color $C$ of a pixel by blending $N$ ordered Gaussians overlapping the pixel :
\[%\begin{equation}
\label{eq:gauss_color}
    C = \sum_{i \in N}c_i\alpha_i \prod_{j=1}^{i-1}(1-\alpha_j).
\]%\end{equation},
where  $c_i$ is the color of each Gaussian and $\alpha_i$ is given by evaluating a
2D Gaussian with covariance multiplied with a
learned per-Gaussian opacity $\sigma$ \cite{Yifan_2019,kerbl3Dgaussians}. This representation allows for fast rendering of novel views, and aims to reconstruct the geometry of the scene.

\subsection{Deformation Fields for Dynamic Scenes}

Prior work showed that a deformation function combined with a static NeRF in a canonical space can enable novel view synthesis in dynamic scenes. %For every point, Equation~\ref{eq:deformation_nerf} shows how
The deformation function $F_\mathrm{NeRF}:\mathbb{R}^3 \rightarrow \mathbb{R}^3$ deforms a point in world coordinates $(x')$ into a point in canonical space $(x)$. 
%\begin{equation}
% \label{eq:deformation_nerf}
%     x',y',z' = F(x,y,z,t)
%\end{equation}

% \JI{Lower-case $\sigma$ is only defined for NeRFs. It needs to be defined for GS.}
Prior work formulated $F_\mathrm{NeRF}$ as an MLP~\cite{dnerf} and a multi-resolution voxel grid~\cite{TiNeuVox}. Wu et al.~\cite{4dgs} applied a similar approach to arrive at Gaussian splatting of dynamic scenes. Given the state of a single canonical Gaussian, defined by $P = [\mu,S,R,\sigma,C]$  at time $t$, a deformation function is%is learned in the form of Equation~\ref{eq:gauss_deformation}.
\[%\begin{equation}
    \label{eq:gauss_deformation}
    P' = F_\mathrm{4DGS}(P,t),
\]%\end{equation}
where $F_\mathrm{4DGS}$, similar to the Hexplanes~\cite{cao2023hexplane}, contains a neural-voxel encoding in space and time. 4D-GS~\cite{4dgs} starts with a \emph{coarse} stage for initializing the canonical space, by setting $P'$ = $P$, bypassing the deformation field and learning canonical properties directly. During the \emph{fine} stage we learn the deformation function. 

\begin{figure}[t]
    \centering
    \includegraphics[width=0.96\linewidth]{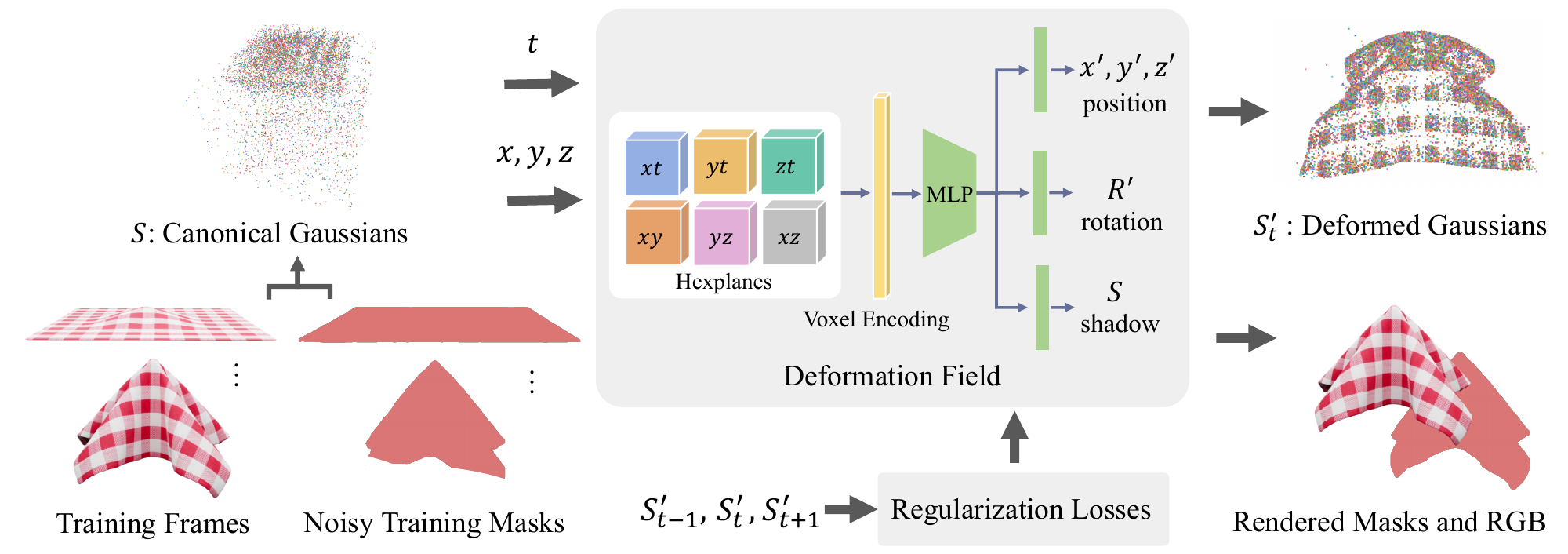}
    \vspace{-2mm}
    \caption{\modelname{} maps a set of Gausians with canonical properties to metric space using a deformation function $F$. The deformation function takes in the position of a Gaussian $(x,y,z)$ and a queried timestamp $t$, to infer shadow $s$, rotation $r'$ and metric position $x'$. During training, we use the metric positions and rotations to regularize the deformation function, considering the state at $t = \{i-1,i,i+1\}$ with Gaussian metric states $P'_{t-1},P'_{t},P'_{t+1}$ }
    \label{fig:4dgs_method}
    \vspace{-0.8em}
\end{figure}

% \JI{avoid words like `simple'} 
We propose \modelname{} (Figure~\ref{fig:4dgs_method}), based on 4D-Gaussians~\cite{4dgs}, to render novel views in dynamic scenes. The key differences with 4D-GS are: (1) we propose an intuitive method to track canonical Gaussians in world coordinates using a continuous deformation function, (2) the output of the deformation function is different, e.g., \modelname{} infers shadows and does not alter opacity or scale over time, and (3) using the method shown in Figure~\ref{fig:canonical_tracking}, we enforce physics-inspired regularization losses on the 3D trajectories of Gaussians.

% \subsection{Dynamic Novel View Synthesis}
\section{Method}
\label{sec:method}
\modelname{} achieves novel-view synthesis and high-quality 3D tracking using a canonical space of Gaussians and a deformation function to deform them to world space (Section~\ref{sec:4d_gs}). To incentivize learning physically plausible deformations, \modelname{} introduces several regularization terms
%We then discuss the regularization terms introduced to incentivize the deformation function to physically plausible deformations
(Section~\ref{sec:regularization}). Finally, \modelname{} learns 3D masks to focus regularization and Gaussian deformation on dynamic parts of the scene (Section~\ref{sec:masking}).

\subsection{4D Gaussian Splatting}
\label{sec:4d_gs}

\textbf{Canonical Neural Voxel Encoding.}
As with prior work, \modelname{} learns a deformation function $F$ from a canonical space. We use a neural-voxel encoding to ensure $F$ has sufficient capacity to capture complex deformations. Prior work~\cite{kplane,4dgs,TiNeuVox,mueller2022instant} showed that neural-voxel encodings improve the speed and accuracy of novel-view synthesis in dynamic scenes. We leverage HexPlanes~\cite{cao2023hexplane,4dgs} to increase capacity for simultaneous 3D tracking and novel-view synthesis. 

% Explain neural voxel encoding
Figure~\ref{fig:4dgs_method} shows an overview of the canonical neural-voxel encoding. Each of the six voxel modules can be defined as  $R(i,j) \in \mathbb{R}^{h \times lN_i \times lN_j}$. Here $\{{i,j}\} \in \{(x,y),(x,z),(y,z),(x,t),(y,t),(z,t)\}$, i.e., we adopt HexPlanes in all possible combinations. $h$ is the size of each feature vector in the voxel, $N_i, N_j$ are the sizes of the HexPlanes in each dimension, $l$ is the upsampling scale. In every module, each plane has a different upsampling scale $l$. To query the multi-resolution voxel grids, we query each plane using bilinear interpolation to finally arrive at a feature vector used by the deformation MLP.

\textbf{Deformation MLP.}
The deformation MLP takes in the voxel encoding and uses the encoding to deform the canonical Gaussians into world coordinates. Figure~\ref{fig:4dgs_method} shows the deformation MLP, which infers position, rotation, and a shadow scalar, given a feature vector from the neural voxel encoding. We choose this set of outputs to model rigid-body transformations of each Gaussian and changes in illumination. Modeling changes in illumination is critical in the presence of shadows. We multiply the RGB color of each Gaussian by the shadow scalar $s \in [0,1]$, and the shadow scalar is in the range $[0,1]$ by feeding the output of the MLP through a sigmoid activation function. 

Next, we deform the Gaussians, modifying their mean positions $\mu$ and rotation $R$, and arrive at a set of Gaussians in the world space each with state $P$. The differentiable rasterizer from Gaussian Splatting~\cite{kerbl3Dgaussians} then renders the Gaussians to retrieve gradients for regressing both the canonical Gaussian states and the parameters of the deformation function.

Unlike 4D-Gaussians~\cite{4dgs}, we propose to not infer opacity or scale using the deformation field. Optimizing for opacity and scale over time would allow Gaussians to disappear or appear instead of following the motion, which would make tracking less accurate. This design choice reduces the capacity of the deformation function, hence a lower view reconstruction quality as compared to 4D-Gaussians might be expected.

\subsection{3D Tracking using 4D Gaussians}
\label{sec:regularization}

\begin{figure}[t]
    \centering
    \includegraphics[width=0.5\linewidth]{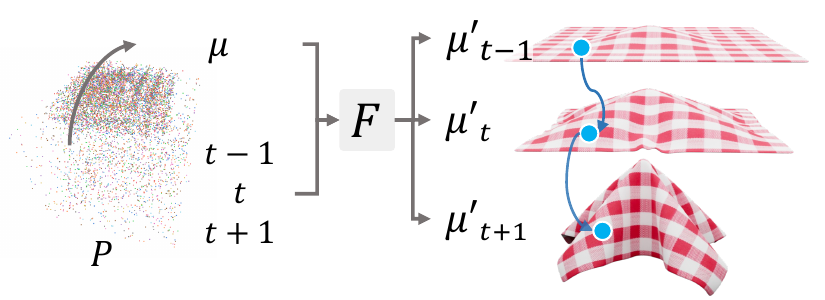}
    \caption{\modelname{} uses three adjacent timesteps at every iteration to enforce physics-inspired regularization terms. All Gaussians are deformed to world space using the deformation function $F$, and rasterized to compute the photometric loss and its gradients. The positions of the Gaussians are used to compute the regularization terms based on local isometry and conservation of momentum (Section~\ref{sec:regularization}).  }
    \label{fig:canonical_tracking}
    \vspace{-0.8em}
\end{figure}
\textbf{Physics-Inspired Losses.} Figure~\ref{fig:canonical_tracking} shows the process of tracking Gaussians from the canonical space in world space. By querying the deformation function $F$, we can track the position of a Gaussian along the entire trajectory.

Without additional supervision, this approach will not necessarily converge to physically plausible deformations. Especially when objects include areas with little texture and uniform color, the solution space for all deformations is underconstrained by photometric consistency alone. To learn a more grounded deformation function, we propose regularization terms inspired by physics.

% It is challenging to design regularization terms for moving 3D Gaussians that represent highly deformable scenes, but also include rigid-body transformations. For example, a cloth might be dropped on the ground (Scene 1), forcing all Gaussians to undergo a sudden deceleration, but otherwise undergo smooth motion. 

After empirically evaluating several combinations of regularization terms, we adopt the isometry loss proposed in~\cite{dyna3dgs} and add a conservation of momentum term. The first term captures a local isometry loss, which we compute based on the state of the $k$ nearest neighboring (KNN) Gaussians. 

\textbf{Local Isometry Loss}
We incentivize the Gaussians to keep the relative position of the $k$ nearest neighbors constant w.r.t. $t=0$. With sufficient deformation, this assumption will be broken at a larger scale, but at a local scale, this regularization avoids drift from the ground-truth trajectory. The isometry loss is
\[%\begin{equation}
% NB: using negative spacing tricks to squeeze this into the margins. \! = a small negative space. {-} removes the space before and after the minus sign (technically, it tells latex that the minus sign is not an operator and thus does not get operator spacing).
   \mathcal{L}_{t}^{\text{iso}} = \frac{1}{k | \mathcal{P} |} \! \sum_{i \in \mathcal{P}}\! \sum_{j \in \text{knn}_{i}}\!\! w_{i,j} \! \left| \left\| \mu_{j,0} {-} \mu_{i,0} \right\|_2 - \left\| \mu_{j,t} {-} \mu_{i,t} \right\|_2  \right|.
\]%\end{equation}
with
\[%\begin{equation}
%\label{eq:weight_reg}
w_{i,j} = \text{exp} \left( - \lambda_w \|\mu_{j,0}-\mu_{i,0} \|_2^2  \right),
\]%\end{equation}
Here $\mathcal{P}$ is the set of all Gaussians.

\textbf{Conservation of Momentum}
We add a term to incentivize conservation of momentum. Newton's first law states objects without external forces applied, given some mass $m$ and velocity vector $\mathbf{v}$, maintain their momentum $m\cdot \mathbf{v}$. We introduce the regularization term %in Equation~\ref{eq:momentum}.
\[%\begin{equation}
\label{eq:momentum}
    \mathcal{L}^{\text{momentum}}_{i,t} = \| \mu_{i,t+1} + \mu_{i,t-1} -2 \mu_{i,t}   \|_1.
\]%\end{equation}
This term incentivizes a constant-velocity vector and has the effect of imposing a low-pass filter on the 3D trajectories. It smooths out trajectories with many sudden changes of direction and magnitude (momentum). 

\subsection{Learning 3D Masks}
\label{sec:masking}
Learning accurate 3D tracking in scenes with a mix of static and dynamic objects and rich textures poses significant challenges, mainly: (1) imposing physics-inspired regularization terms on all Gaussians may cause issues when dynamic and static objects interact, and (2) modeling millions of dynamic Gaussians can become a significant computational burden.

To address this, \modelname{} takes noisy masks of dynamic scene components such as cloth, and learns what Gaussians are dynamic. More formally, we render a mask $M$ by
\[%\begin{equation}
\label{eq:gauss_color}
    M = \sum_{i \in N}m_i\alpha_i \prod_{j=1}^{i-1}(1-\alpha_j).
\]%\end{equation},
where $m_i$ is a per-Gaussian property, with $m_i \in [0,1]$. We then add a regularization term to the loss function s.t. $m_i$ is regressed to best reconstruct M. Finally, \modelname{} uses $m_i$
to select a subset of Gaussians to be dynamic, and applies regularization terms only to those Gaussians.

% Table~\ref{tab:track} shows the impact of the proposed regularization terms and other design choices on tracking errors.

\section{Experiments}
\label{sec:experiments}
We evaluate \modelname{} on synthetic and real-world datasets of scenes with highly deformable objects. Section~\ref{sec:sim_exp_setup} provides details on the simulation experiment setup, evaluation metrics, and baseline methods. Section~\ref{sec:sim_exp_result} reports evaluation results from the compared methods and provides analysis. Section~\ref{sec:real_setup} lists the real-world evaluation setup, and finally in Section~\ref{sec:real_results} we provide a qualitative evaluation of the performance of \modelname{} in the real-world.

% Section \ref{exp_ablate} supplies additional ablation results on the training efficiency of \modelname{}.  

\subsection{Simulation Experiment Setup}
\label{sec:sim_exp_setup}

\textbf{Dataset Preparation} We use Blender to model dynamic cloth sequences and render photo-realistic images. We create 6 distinct scenes, each containing a different cloth with distinct visual and physical properties, and we render images from 100 different camera views and 40 consecutive time steps for training for a total of 4,000 images. The cloth deformations are introduced by dropping each cloth over one or a few invisible balls onto a ground plane or by constraining the cloth at an attachment point. We obtain ground-truth trajectories by tracking the mesh vertices of deformable objects in Blender. Every scene contains a single deformable object and a rendered background.
  
\textbf{Oracle Baselines} We compare \modelname{} to 2D tracking oracle models which have access to ground truth depth and trajectory information. While these methods were not designed for 3D tracking, they are well-known for their impressive 2D tracking performance. Their numbers aid in putting the tracking performance of the other baselines into context. We run RAFT~\cite{teed2020raft} on all views, project tracking to 3D using ground truth depth, and report the mean results as the RAFT model. We provide two additional oracle methods which have access to the ground-truth trajectories as well. \emph{RAFT Oracle} first evaluates on all views, to then output only the trajectories from the view with the lowest median trajectory error. We also report \emph{OmniMotion Oracle}, which runs OmniMotion~\cite{trackeverything} on the viewpoint with the lowest MTE for RAFT. Training OmniMotion takes roughly 12--13 hours on an Nvidia RTX 4090 GPU, making inference on all 100 views impractical. The numbers from \emph{RAFT Oracle} and \emph{OmniMotion Oracle} are not an apples-to-apples comparison with the other methods, as to obtain their result they had to access privileged ground-truth trajectories. 

% (2) DeVRF~\cite{liu2022devrf} \BD{only add if jiawei can run this} represents a dynamic, non-rigid scene with a 3D canonical voxel space and a 4D deformation voxel field. It proposes a 4D deformation cycle that models both the backward deformation from the deformed space to the canonical space and the forward deformation from the canonical space to the deformed space. Therefore, the point trajectories can be obtained by querying the forward deformation of canonical points at $t=0$ to every timestep's deformed space through the 4D forward deformation space.  While DeVRF was not designed for 3D tracking directly, it provides useful insight into how similar methods based on NeRF might perform in highly deformable scenes.

\textbf{Gaussian Splatting Baselines} (1) Dynamic 3D Gaussians (\emph{DynaGS})~\cite{dyna3dgs}, which also builds on 3D Gaussian splatting for dynamic novel-view synthesis, except it explicitly models the positions and rotations of each Gaussian at each time-step. This results in straightforward tracking of any point via finding the trajectory of the learned Gaussian closest to a queried point. Although the original paper assumes a known point cloud at the first frame, we instead use a randomly sampled point cloud for a fair comparison, with DynaGS and \modelname{} both not using depth information. 

(2) Finally, we compare to tracking using 4D-Gaussians~\cite{4dgs} (\emph{4D-GS}). We add the approach for 3D tracking of canonical Gaussian, as shown in Figure~\ref{fig:canonical_tracking}, to extract 3D trajectories from a learning view synthesis model. Comparing to \emph{4D-GS} serves to show the impact of the changes made in the model architecture, the regularization terms, and using learning per-Gaussian masks to arrive at \modelname{}.

\textbf{Training and Evaluation Setup}
We create a dataset of 6 dynamic cloth scenes, each with varying physical and visual properties (Figure~\ref{fig:teaser}). For \modelname{} and 4D Gaussians, we perform 30,000 training iterations, and set point cloud pruning interval to 100, voxel plane resolution to [64, 64], and multi-resolution upsampling to levels $L = \{1, 2, 4, 8\}$. We set the regularization hyper parameters (Section~\ref{sec:regularization}) for all synthetic scenes to $\lambda_w = \text{2,000}$, $\lambda^{\text{momentum}} = \text{0.03}$, $\lambda^{\text{iso}} = \text{0.3}$, and $k = \text{20}$ for KNN. We keep all hyper parameters the same for real-world scenes, but increase the regularization terms for momentum and isometry loss. We generate the masks with segment anything (SAM)~\cite{kirillov2023segment} for the initial frame, and use XMem~\cite{cheng2022xmem} to propagate to future frames.

For DynaGS, we set $\lambda^{\text{rigid}} = \text{4}$, $\lambda_w = \text{2,000}$,$\lambda^{\text{iso}} = \text{2.0}$,  and $k = 20$, as in the open-source code. 

We evaluate each compared method on 1,000 randomly sampled points on each cloth.

    % \vspace{-5mm}
    
% \begin{figure*}[t]
%     \centering
%     \includegraphics[width=0.9\linewidth]{Figures/fig5.pdf}
%     \caption{Qualitative novel view synthesis results from all baselines and \modelname{}. Both 4D-GS~\cite{4dgs} and \modelname{} achieve the best qualitative results that are close to ground-truth renderings (especially on Scene 6, where the floral patterns on the cloth are reconstructed), albeit they still fail to capture the fine-grained texture and wrinkles on Scene 2 and Scene3 cloth.}
%     \label{fig:result-view}
%     \vspace{-0.8em}
% \end{figure*}

\subsection{Simulation Results}
\label{sec:sim_exp_result}

\begin{table}[t] 
\centering
% \scriptsize
\resizebox{\textwidth}{!}{%
\begin{tabular}{llrrrrrrr}
\toprule
 Metric                     & Method      &   Scene 1 &   Scene 2 &   Scene 3 &   Scene 4 &   Scene 5 &   Scene 6 & Mean \\
\hline
 
\multirow{6}{*}{3D MTE [mm]~$\downarrow$} & RAFT~\cite{teed2020raft}  \textbf{$^a$} &   67.264  &  89.944  & 220.125   & 177.909    &  84.593   &  23.422  & 110.543  \\
                                     & RAFT Oracle~\cite{teed2020raft}\textbf{$^a$}\textbf{$^b$} &    3.381 &    26.956 &    58.971 &    12.481 &     3.930 &     3.192 & 18.152 \\
                                     & OmniMotion Oracle~\cite{trackeverything}  \textbf{$^a$}\textbf{$^b$}     &   0.535 &    14.513 &    39.958 &     4.556 &     2.487 &     2.011    & 10.677  \\
                                     \cmidrule{2-9}
                                     
                                     & DynaGS~\cite{dyna3dgs}        &24.233 &    81.119 &   464.64  &    54.074 &    34.985 &    36.101     &  115.859 \\
                                     & 4D-GS~\cite{4dgs}           &     4.645 &    95.032 &   223.839 &    27.441 &    14.091 &    11.619 &  62.778 \\
                                     & \bf \modelname{}          &  \textbf{3.373} &    \textbf{45.33}  &    \textbf{88.369} &    \textbf{14.022} &     \textbf{7.257} &     \textbf{8.173}  & \textbf{27.754} \tabularnewline

                                     \cmidrule{1-9}
\multirow{6}{*}{3D~$\delta_\mathrm{avg}~\uparrow$}              & RAFT~\cite{teed2020raft}  \textbf{$^a$} &  0.553   &  0.379  &  0.222  &  0.533   &  0.586   & 0.619   & 0.482 \\
                                     & RAFT Oracle~\cite{teed2020raft}\textbf{$^a$}\textbf{$^b$} &    0.926 &     0.577 &     0.411 &     0.703 &     0.833 &     0.808  & 0.710\\
                                     & OmniMotion Oracle~\cite{trackeverything}  \textbf{$^a$}\textbf{$^b$}     &    0.987 &     0.693 &     0.549 &     0.864 &     0.885  &     0.849 & 0.805  \\
                                     \cmidrule{2-9}
                                     
                                     & DynaGS~\cite{dyna3dgs}        &    0.624 &     0.31  &     0.042 &     0.435 &     0.533 &     0.527 & 0.4118\\
                                     & 4D-GS~\cite{4dgs}           &      0.902 &     0.339 &     0.164 &     0.583 &     0.697 &     0.715&  0.5667 \\
                                     & \bf \modelname{}          &   \textbf{0.929} &     \textbf{0.522} &     \textbf{0.322} &     \textbf{0.71}  &   \textbf{0.856} &     \textbf{0.787}  &  \textbf{0.688}  \tabularnewline
\cmidrule{1-9}
\multirow{6}{*}{3D Survival~$\uparrow$}     &  RAFT~\cite{teed2020raft}  \textbf{$^a$} &  0.945   & 0.792   & 0.779   & 0.822    &  0.792   &  0.854  &  0.831 \\
                                     & RAFT Oracle~\cite{teed2020raft}\textbf{$^a$}\textbf{$^b$} &  0.986 &     0.833 &     0.957 &     0.872 &     0.929 &     0.903     & 0.913 \\
                                     & OmniMotion Oracle~\cite{trackeverything}  \textbf{$^a$}\textbf{$^b$}     &  1     &     0.963 &     1     &     0.985 &     0.963 &     0.933     & 0.977  \\
                                     \cmidrule{2-9}
                                     
                                     & DynaGS~\cite{dyna3dgs}        &   0.999 &     0.99  &     0.483 &     0.988 &     0.992 &     0.992  & 0.907  \\
                                     & 4D-GS~\cite{4dgs}           &      \textbf{1}     &     0.967 &     0.834 &     \textbf{1}     &     0.999 &     \textbf{1} & 0.967   \\
                                     & \bf \modelname{}          &    \textbf{1}     &     \textbf{1}     &     \textbf{1}     &     \textbf{1}     &     \textbf{1}     &     \textbf{1} & \textbf{1} \tabularnewline
\bottomrule
\multicolumn{9}{l}{\scriptsize \textbf{$^a$} Method had access to ground truth depth. \textbf{$^b$} Method had access to ground truth trajectories to pick the best camera view.}
% \multicolumn{8}{l}{\small \textbf{$^b$} Method had access to ground truth depth .}
\end{tabular}}
\caption{3D tracking results on the deformable cloth dataset (Figure~\ref{fig:teaser}). For each metric, the methods above the solid line had access to privileged information, see \textbf{$^a$}\textbf{$^b$} and Section~\ref{sec:sim_exp_setup} for more details. The results suggest that \modelname{} outperforms the baselines in all averaged metrics, and is competitive with the oracle models. The results also suggest our novel deformation function architecture, learning per-Gaussian masks, and physics-inspired regularization losses improve the tracking performance compared to 4D-GS~\cite{4dgs}. We do not consider the oracle methods to be fair baselines and therefore do not bold their results.   }
\label{tab:track}
\vspace{-0.8 em}
\end{table}

\textbf{3D Point Tracking} Following prior work \cite{pointodyssey,dyna3dgs}, we report median trajectory error (MTE), position accuracy ($\delta$), and
the survival rate with a threshold of 0.5 [m]~\cite{dyna3dgs}. 

The results are summarized in Table \ref{tab:track}. We make the following observations: (1)~\modelname{} outperforms baselines RAFT, DynaGS, and 4D-GS, by achieving a MTE of 55.8\% - 76.0\% lower compared to the baselines. (2) The discrepancy between the RAFT oracle model and its averaged result demonstrates the difficulty arising from frequent self-occlusions. This also points to future research avenues for additional supervision through optic flow and 2D tracking algorithms such as RAFT. (3) The oracle models perform very well, this is in part thanks to the falling and short-horizon nature of these sequences, limiting self-occlusions. In the real-world we expect much larger errors due to noisy depth and more challenging occlusions in long-horizon tasks. It would also be unclear what viewpoint to choose without access to ground truth trajectories. (4) Scenes with less texture such as scene 3 perform significantly worse than scenes with strong texture.

\begin{figure}[htb!]
        \centering
        \begin{subfigure}[b]{0.24\linewidth}
            \centering
            \includegraphics[width=\linewidth]{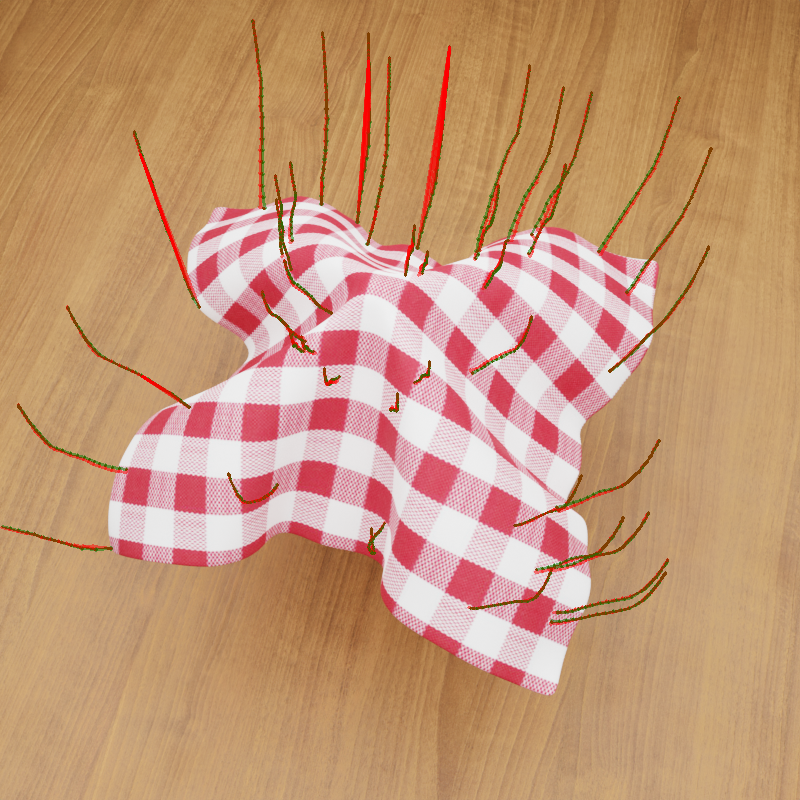}
            \caption[Network2]%
            {{\small RAFT~\cite{teed2020raft} }}    
            \label{fig:devrf_quant}
        \end{subfigure}
        \hfill
        \begin{subfigure}[b]{0.24\linewidth}  
            \centering 
            \includegraphics[width=\linewidth]{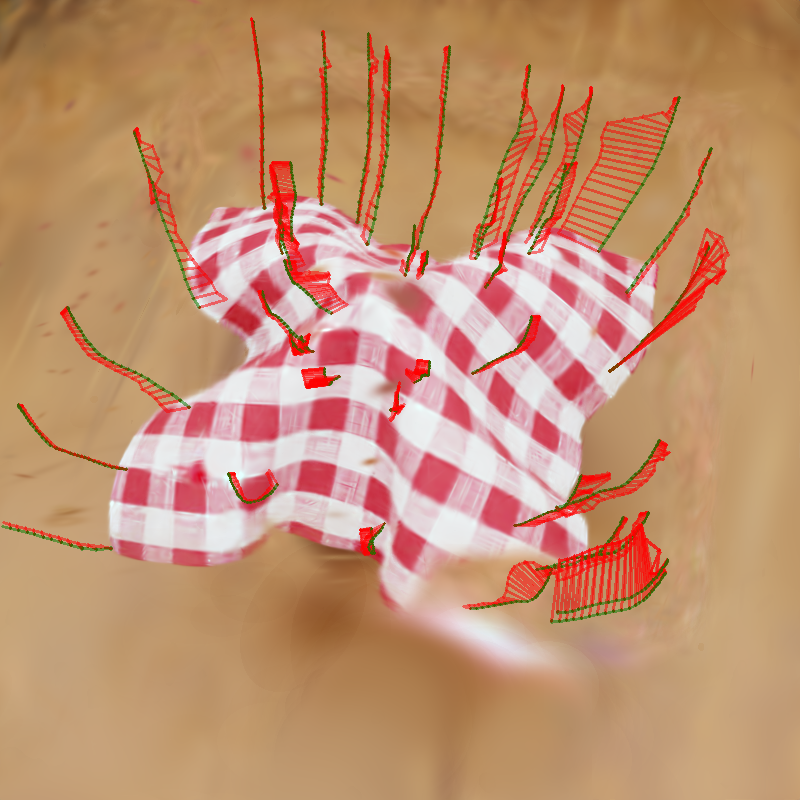}
            \caption[]%
            {{\small DynaGS~\cite{dyna3dgs}}}    
            \label{fig:dyna_gs_quant}
        \end{subfigure}
        \begin{subfigure}[b]{0.24\linewidth}   
            \centering 
            \includegraphics[width=\linewidth]{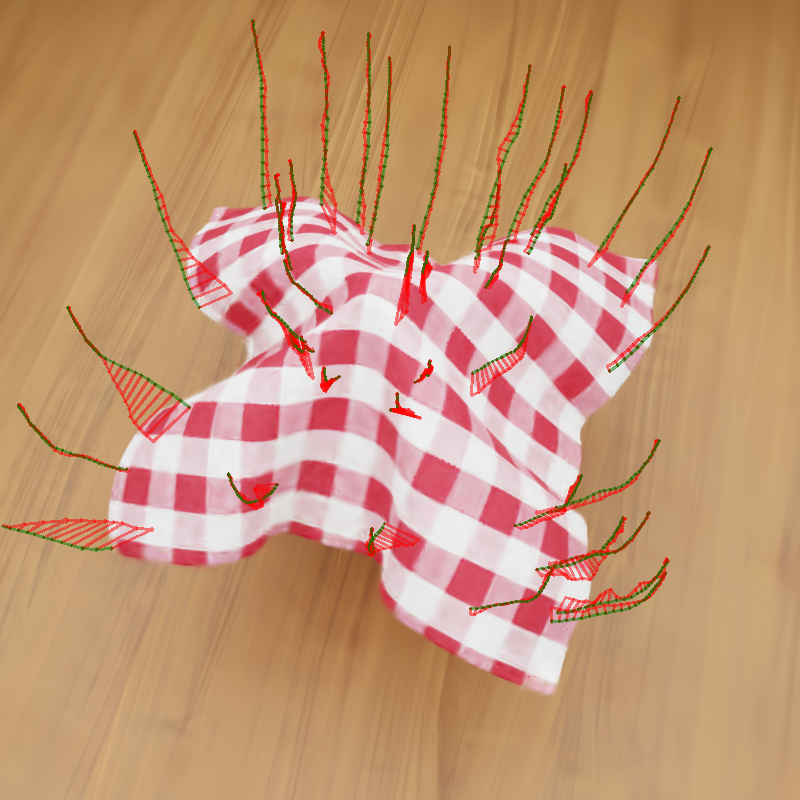}
            \caption[]%
            {{\small 4D-GS~\cite{4dgs}}}    
            \label{fig:4dgs_quant}
        \end{subfigure}
        \hfill
        \begin{subfigure}[b]{0.24\linewidth}   
            \centering 
            \includegraphics[width=\linewidth]{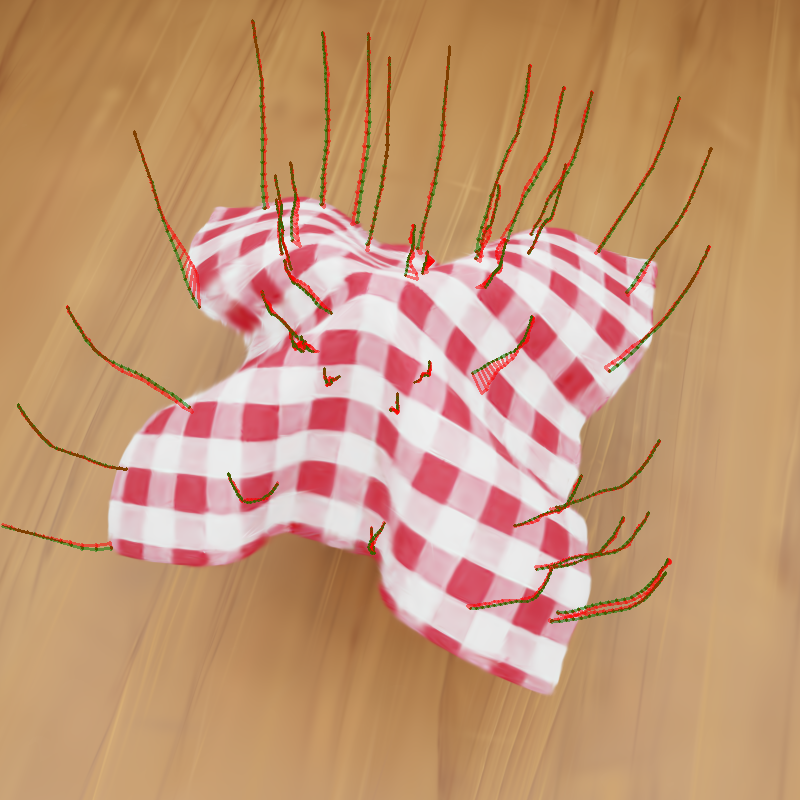}
            \caption[]%
            {{\small \modelname{}}}    
            \label{fig:mdsplat_quant}
        \end{subfigure}
        \caption[]
        {\small \textbf{Results on Scene 5}: randomly sampled ground-truth trajectories in green, inferred trajectories in red, and the error of corresponding points in red lines. Compared to the baseline methods, \modelname{} results in fewer errors in 3D tracking.} 
        \label{fig:trajs_qual}
    \vspace{-0.8 em}
\end{figure}

\textbf{Qualitative Results} 
Figure~\ref{fig:trajs_qual} shows ground truth and inferred trajectories in scene 5. The results show that especially DynaGS and 4D-GS introduce large errors as the cloth drapes down. RAFT improves over DynaGS and 4D-GS but requires accurate depth estimation.

\subsection{Real-World Experiment Setup}
\label{sec:real_setup}
\textbf{Robo360 Data} The Robo360 dataset~\cite{liang2023robo360} is a 3D omnispective multi-material robotic manipulation dataset. It covers many different scenario's, including manipulation by robot manipulators and humans captured by 86 calibrated cameras. These properties make it an ideal dataset to evaluate the effectiveness of \modelname{} in the real world.  We select two scenes: (1) a human folding a larger duvet and (2) a human folding a smaller cloth. In the cloth folding scene, we exclude viewpoints where the entire person's body is visible to eliminate unnecessary complexity. 

We also subsample the data to demonstrate \modelname{} performance with fewer views. The duvet folding scene contains 17 training views and the cloth folding scene contains 20 training views.

% \textbf{A low-cost sparse camera setup} \BD{remove if not successful} While the Robo360 dataset contains high-quality captures from many viewpoints, the large number of DSLRs used in the setup might limit the scalability of the approach. We therefore also test \modelname{} on a dataset with 16 views. Each camera is mounted to a tripod and connected to a portable computer to save the recordings. All cameras are started simultaneously by a host computer connected to all portable nodes.

\subsection{Real-World Experiment Results}
\label{sec:real_results}
\begin{figure}[t]
    \centering
    \includegraphics[width=\linewidth]{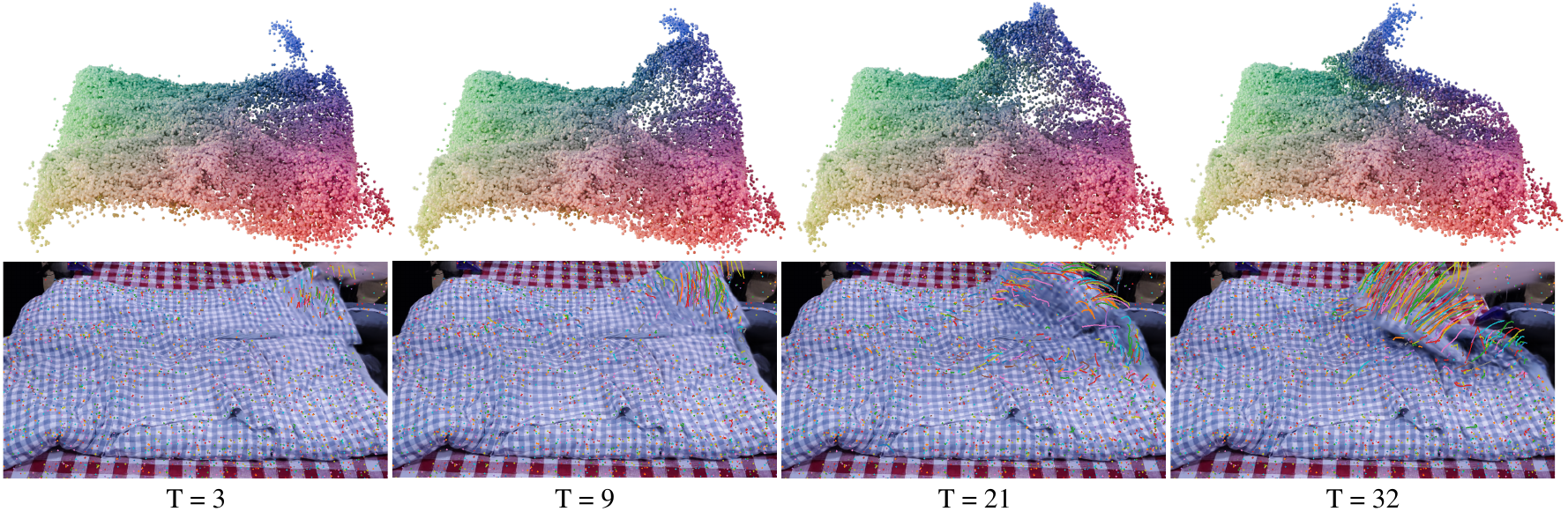}
    \caption{A person manipulating a duvet in the Robo360~\cite{liang2023robo360} dataset, reconstructed using \modelname{}. The top row shows the 4D Gaussians as point clouds, where the color represents dense correspondences. The bottom row shows rendered views overlaid with 3D trajectories projected to image space. }
    \label{fig:duvet_points}
\end{figure}

\textbf{Real2Sim for Digital Twins} Figure~\ref{fig:duvet_points} shows the 3D tracking overlaid on rendered images, as well as the Gaussian points at each time step. The results suggest that \modelname{} is able to successfully infer smooth and meaningful trajectories in the real world. While no ground truth is available, the trajectories appear to follow their geometry closely except for a few floating Gaussians. Hyperparameter tuning of the regularization functions, as well as discarding Gaussians with a low opacity, might help resolve this.

The point cloud included in this Figure can be used to create a digital twin after recording the sequence. The digital twin of the duvet, and the entire environment, can then be used to create more dense supervision for imitation learning approaches. 

\begin{figure}[htb!]
        \centering
            \includegraphics[width=\linewidth]{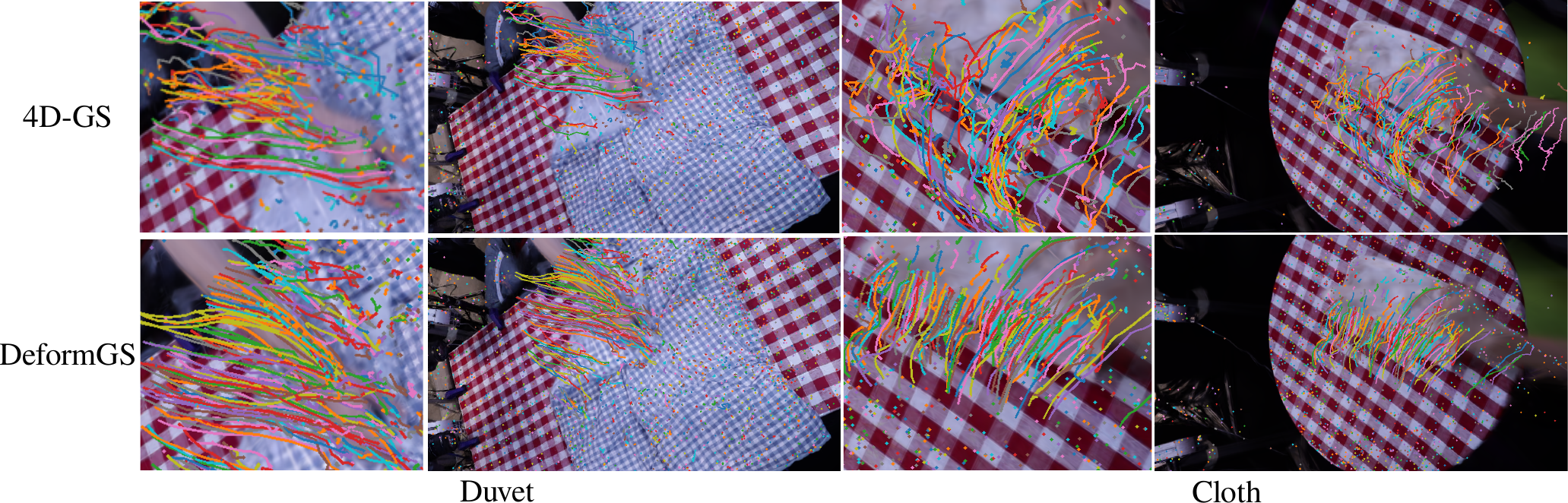}
            \caption[Network2]%
            {{\small Real-world results comparing our proposed \modelname{} against 4D-GS~\cite{4dgs}. The 3D trajectories inferred by \modelname{} appear more smooth and accurate, whereas 4D-GS displays more cluttered trajectories.}}    
            \label{fig:both_comparison}
    \vspace{-0.8 em}
\end{figure}

\textbf{Task-Relevant Keypoint Tracking} Robotic manipulators can benefit from tracking task-relevant keypoints, such as the corner of a cloth or the edge of a jacket. Figure~\ref{fig:both_comparison} shows a comparison between 4D-GS~\cite{4dgs} and \modelname{} in 3D point tracking, evaluated on both duvet and cloth scenes. The results suggest \modelname{} leads to more smooth and overall useful trajectories. The trajectories from 4D-GS intertwine into more messy trajectories, and appear less physically plausible. This would hinder the adoption of 3D tracking into robot applications.

% \begin{figure}[t]
% \centering
% \includegraphics[width=\linewidth]{Figures/duvet_comparison.pdf}
% \caption{4D-GS compared against \modelname{} on a duvet folding task. The result suggests \modelname{} infers smooth and more accurate tracking, as opposed to the more cluttered result from 4D-GS.}
% \label{fig:duvet_comparison}
% \end{figure}

\section{Conclusions}

In this work, we address the challenging problem of 3D point-tracking in dynamic scenes with deformable objects. We introduced \modelname{}, the first approach that learns continuous deformations for 3D tracking of deformable scenes. We empirically demonstrate that \modelname{} outperforms baseline methods and achieves both high-quality dynamic scene reconstruction and high-accuracy 3D tracking on highly deformed cloth objects with occlusions and shadows, both in simulation and the real world. We also contribute a dataset of six synthetic scenes to facilitate future research.

\textbf{Limitations and Future Work} \modelname{}, similar to prior work on dynamic novel view reconstruction, requires a setup of multiple synchronized and calibrated cameras, which may require a significant engineering effort in real-world scenarios. Additionally, significant innovation will be required to achieve the demonstrated results in real-time, as will be beneficial for scalable robot applications. 

While \modelname{} improves upon prior methods, we do observe Gaussians wandering off in some cases. We also notice the algorithm is relatively sensitive to the regularization hyper parameters ($\lambda^{\text{momentum}}$ and $\lambda^{\text{iso}}$), this might be resolved in the future by adding supervision from state-of-the-art point-tracking algorithms. These limitations point to promising directions for future research.

\textbf{Acknowledgements} This work was supported by the Center for Machine Learning and Health (CMLH) at CMU, and the Pittsburgh Super Computing Center (PSC). We thank David Held for the productive discussions.

\begin{credits}
% \subsubsection{\ackname} A bold run-in heading in small font size at the end of the paper is
% used for general acknowledgments, for example: This study was funded

\subsubsection{\discintname}
The authors have no competing interests to declare
that are relevant to the content of this article. 
% It is now necessary to declare any competing interests or to specifically
% state that the authors have no competing interests. Please place the
% statement with a bold run-in heading in small font size beneath the
% (optional) acknowledgments\footnote{If EquinOCS, our proceedings submission
% system, is used, then the disclaimer can be provided directly in the system.},
% for example: The authors have no competing interests to declare that are
% relevant to the content of this article. Or: Author A has received research
% grants from Company W. Author B has received a speaker honorarium from
% Company X and owns stock in Company Y. Author C is a member of committee Z.
\end{credits}
\bibliographystyle{splncs04}
\bibliography{main}

\end{document}